\documentclass[sigconf]{acmart}
\AtBeginDocument{%
  \providecommand\BibTeX{{%
    \normalfont B\kern-0.5em{\scshape i\kern-0.25em b}\kern-0.8em\TeX}}}

\setcopyright{acmcopyright}
\copyrightyear{2023}
\acmYear{2023}
\acmDOI{XXXXXXX.XXXXXXX}
\acmConference[WWW'23]{Make sure to enter the correct
  conference title from your rights confirmation emai}{April 30--May 4,
  2023}{Austin, USA}
\acmPrice{15.00}
\acmISBN{978-1-4503-XXXX-X/18/06}
\usepackage{graphicx}
\usepackage{amsmath}
\usepackage{amssymb}
\usepackage{booktabs}
\usepackage{multirow}
\usepackage{subfigure}
\usepackage{flushend}
\usepackage{balance}

\setcopyright{none}
\renewcommand\footnotetextcopyrightpermission[1]{} % removes footnote with conference information in first column
\begin{document}
%%
%% The "title" command has an optional parameter,
%% allowing the author to define a "short title" to be used in page headers.
\title{DEDGAT: Dual Embedding of Directed Graph Attention Networks for Detecting Financial Risk}
%\title{Bidirectional graph attention networks for detecting financial risks}
%%
%% The "author" command and its associated commands are used to define
%% the authors and their affiliations.
%% Of note is the shared affiliation of the first two authors, and the
%% "authornote" and "authornotemark" commands
%% used to denote shared contribution to the research.
\author{Jiafu Wu$^*$, Mufeng Yao}
%\author{Jiafu Wu$^\authornote $, Mufeng Yao}
% \author{Jiafu Wu\footnotemark[1], Mufeng Yao\footnotemark[1]}
\authornote{These authors contributed equally to this work and should be considered co-first authors.}
% \authornote{These authors contributed equally to this work and should be considered co-first authors.}
% \author{Dong Wu, Mingmin Chi\footnotemark[2]}
\author{Dong Wu, Mingmin Chi}
\authornote{Corresponding authors (mmchi@fudan.edu.cn, yike.wbk@antgroup.com).}
% \authornote{Corresponding authors (mmchi@fudan.edu.cn, yike.wbk@antgroup.com).}
%\thanks{\\$*$& These authors contributed equally to this work and should be considered co-first authors.\\
%$*$ Corresponding author (mmchi@fudan.edu.cn).}
%\authornote{Both authors contributed equally to this research.}
%\authornote{*Corresponding authors}
%\email{\{wujf21,yaomf21,dongwu19\}@m.fudan.edu.cn}
%\orcid{1234-5678-9012}
%\author{Mufeng Yao}
%\authornotemark[1]
%\email{21110240091@m.fudan.edu.cn}
\affiliation{%
  \institution{Fudan University}
  %\streetaddress{No. 2005, Songhu Road}
  \city{Shanghai}
  \country{China}
  %\postcode{43017-6221}
}
\author{Baokun Wang$\dagger$, Ruofan Wu, Xin Fu \\Changhua Meng, Weiqiang Wang}
% \author{Baokun Wang\footnotemark[2], Ruofan Wu, Xin Fu \\Changhua Meng, Weiqiang Wang}
\affiliation{%
 \institution{Tiansuan Lab,Ant Group}
  \city{Shanghai}
  \country{China}}
%\email{\{yike.wbk, ruofan.wrf, zicai.fx, changhua.mch, weiqiang.wwq\}@antgroup.com}
%%
%% By default, the full list of authors will be used in the page
%% headers. Often, this list is too long, and will overlap
%% other information printed in the page headers. This command allows
%% the author to define a more concise list
%% of authors' names for this purpose.
\renewcommand{\shortauthors}{Wu and Yao, et al.}

% \renewcommand{\thefootnote}{\fnsymbol{footnote}} %将脚注符号设置为fnsymbol类型，即特殊符号表示
% \footnotetext[1]{These authors contributed equally to this work.} %对应脚注[1]
% \footnotetext[2]{Corresponding authors.} %对应脚注[2]
%%
%% The abstract is a short summary of the work to be presented in the
%% article.
%摘要增加,在实际应用当中图可能是有向的
%买卖转账与入度出度的因果关系加强
%金融风控方向性的重要性
%与其他有向图的对比
%实际生产当中的应用,成功应用在了实际风控场景中。
\begin{abstract}
Graph representation plays an important role in the field of financial risk control, where the relationship among users can be constructed in a graph manner. In practical scenarios, the relationships between nodes in risk control tasks are bidirectional, e.g., merchants having both revenue and expense behaviors. Graph neural networks designed for undirected graphs usually aggregate discriminative node or edge representations with an attention strategy, but cannot fully exploit the out-degree information when used for the tasks built on directed graph, which leads to the problem of a directional bias. To tackle this problem, we propose a \textbf{D}irec-ted \textbf{G}raph \textbf{AT}tention network called DGAT, which explicitly takes out-degree into attention calculation. In addition to having directional requirements, the same node might have different representations of its input and output, and thus we further propose a dual embedding of DGAT, referred to as DEDGAT. Specifically, DEDGAT assigns in-degree and out-degree representations to each node and uses these two embeddings to calculate the attention weights of in-degree and out-degree nodes, respectively. Experiments performed on the benchmark datasets show that DGAT and DEDGAT obtain better classification performance compared to undirected GAT. Also. the visualization results demonstrate that our methods can fully use both in-degree and out-degree information.

\end{abstract}
%%
%% The code below is generated by the tool at http://dl.acm.org/ccs.cfm.
%% Please copy and paste the code instead of the example below.
%%
%%
%% Keywords. The author(s) should pick words that accurately describe
%% the work being presented. Separate the keywords with commas.

%% A "teaser" image appears between the author and affiliation
%% information and the body of the document, and typically spans the
%% page.
% \begin{teaserfigure}
%   \includegraphics[width=\textwidth]{sampleteaser}
%   \caption{Seattle Mariners at Spring Training, 2010.}
%   \Description{Enjoying the baseball game from the third-base
%   seats. Ichiro Suzuki preparing to bat.}
%   \label{fig:teaser}
% \end{teaserfigure}
%%
%% This command processes the author and affiliation and title
%% information and builds the first part of the formatted document.
\maketitle
\begin{figure}[htb]
 {\centering
 \includegraphics[width=0.40\textwidth]{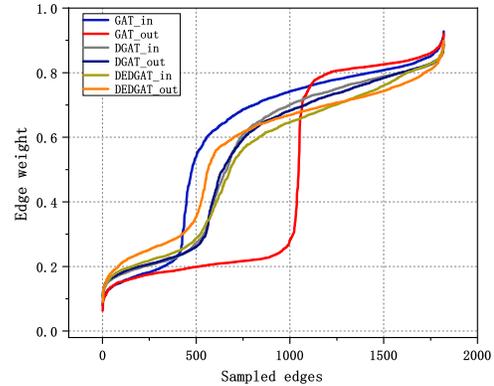}}
	\caption{Importance score distribution of in-degree and out-degree edge weights in a financial risk control dataset with our proposed methods DGAT and DEDGAT compared to GAT on a directed graph. The out-degree edges are far less important than the in-degree ones for GAT. This is defined as a directional bias. Our methods can solve this problem and aggregate in-degree and out-degree represetations in a balanced manner.}
	\label{fig:weight}
\end{figure}
\section{Introduction}
\label{sec:intro}
% 金融风险控制是互联网融资的核心,也是人工智能的主要应用之一\cite{li2009network,aziz2019machine,ma2019financial,yang2020construction}。
Financial risk control is the core of Internet financing and one of the main applications of artificial intelligence~\cite{li2009network,aziz2019machine,ma2019financial,yang2020construction}.
% 在风控场景中,实体之间往往存在各种关系,例如转账关系网、设备关系
% 网、区域关系网等等。
In risk control scenarios, there are often various relationships between entities, such as transfer relationship, device relationship, regional connection, etc.
% 图数据结构将金融数据中各个实体视为图上的节点,将实体之间的关系视为节点之间的边。
% The graph data structure treats each entity in financial data as a node on the graph, and regards the relationship between entities as an edge between nodes.
% 针对图结构的信息提取与挖掘能够帮助我们发现金融系统中的异常对象。
% Extracting information from graph structure can help us discover abnormal objects in the financial system.
Using graph structure to modeling the relationship can help us discover abnormal objects in the financial system.

% 图数据挖掘方法大体上可以分为浅层图表征方法和图神经网络方法。
Graph-based methods can often be divided into shallow graph methods and graph neural network (GNN) methods.
% 以DeepWalk\cite{perozzi2014deepwalk},node2vec\cite{grover2016node2vec},LINE\cite{tang2015line}为代表的浅层图表征仅能够学习图的浅层信息,因此分类性能受限。
The shallow methods represented by DeepWalk~\cite{perozzi2014deepwalk}, node2vec~\cite{grover2016node2vec}, LINE~\cite{tang2015line} can only learn the shallow represetation information of the graph, so the classification performance is limited.
% 与浅层图表征方法不同,图神经网络能够以有监督的方式自动学习图的高阶特征,且能够更充分地利用图上节点与边的特征来辅助节点分类等下游任务,在图节点分类任务上强于浅层方法\cite{weber2019anti}。
Different from shallow methods, GNN can automatically learn high-level features of graphs in a supervised manner and utilize the features of nodes and edges to improve the performance of downstream tasks such as node or edge classification.
% 图卷积网络(GCN)\cite{kipf2016semi}使用卷积来完成节点信息的汇聚。
% GraphSage\cite{hamilton2017inductive}采用邻居采样来汇聚节点的邻域信息,使得大规模图数据的训练成为可能。
Graph Convolutional Network (GCN)~\cite{kipf2016semi} uses convolution to aggregate node information.
GraphSage~\cite{hamilton2017inductive} uses neighbor sampling to aggregate the neighborhood information of nodes, making training on large-scale graph data possible.
% 图注意力网络(GAT)\cite{velivckovic2017graph}将每个节点和它的邻居节点的信息按照注意力进行加权汇聚,广泛地使用在图节点分类任务上。
Graph Attention Network (GAT)~\cite{velivckovic2017graph} aggregates the information of each node and its neighbor nodes according to the weighted attention, which is widely used in the task of graph node classification.
While these methods achieve a better performance compared to shallow methods~\cite{weber2019anti},
% 然而现有的图神经网络大多是针对无向图,因此未能充分挖掘图的方向信息。
most of them are designed for undirected graphs. Accordingly, the direction information of the graph cannot be fully exploited.
% 针对图的有向性,DGCN\cite{ma2019spectral}通过引入一阶近邻和二阶近邻概念将图卷积从无向图扩展到有向图上,\cite{zhang2021}采用Hermitian
% magnetic Laplacian对图的方向性进行解耦。
DGCN~\cite{ma2019spectral} extends graph convolution from undirected graphs to directed graphs by introducing the concepts of 1-st neighbors and 2-nd neighbors to compute Lapalacian matrix. MagNet~\cite{zhang2021magnet} adopts Hermitian magnetic Laplacian to decouple the direction of the graph.
% 这些方法针对有向图进行了优化,但并没有充分考虑入度和出度信息的区分。
These methods are explicitly represented for directed graphs in a transductive way, but do not fully consider the distinction between in-degree and out-degree information.
% 在金融风控领域,入度出度信息的区分是十分重要的,例如银行账户A向账户B转账与B向A转账传递的信息是不同的。此外,虽然A到B的转账是单向的,但是转账关系暗示了A和B之间隐含着双向的社交关系。

In financial risk control, the distinction between in-degree and out-degree information is very important. For example, the information transmitted from bank account A to account B and B to A are significantly different. In addition, although the transfer from A to B is one-way, the transfer relationship implies a two-way social relationship between A and B.
% 因此,针对金融风控的图网络不但应该很好地区分入度和出度信息,也应该对于单向的边采取双向的信息汇聚。
Therefore, the graph network for financial risk control should explicitly distinguish in-degree and out-degree information, and adequately aggregate their information in a balanced manner.
% 基于以上考虑,本文提出有向图注意力网络DA和有向双表征图注意力网络DADE,在对入度出度信息进行区分的基础上充分考虑了双向关系的建模。
However, GAT has a strong directinal bias as shown in Fig.~\ref{fig:weight}, where the out-degree weights are far less important than the in-degree ones. This limits the performance on financial risk control tasks.

To tackle this problem, we propose a directed graph attention network (called DGAT), which calculates the in-degree and out-degree information separately during the feature aggregation process. To better distinguish the in/out-degree represetation on directed graph, we further propose a dual embedding of directed attention network (referred to as DEDGAT) to sovle the directional bias. Specifially, a graph node is divided to in-degree and out-degree nodes respectively and then graph representation is sepratedly computed for in/out-degree nodes. The proposed models achieve SOTA performance on three financial risk control datasets by reducing directional bias verified by GNNExplainer~\cite{ying2019gnnexplainer}.

\section{Proposed Models}

\begin{figure}[htbp]
  \centering
    \subfigure[GAT] {
    	%\begin{minipage}[t]{0.17\textwidth}
    	%\centering          %子图居中
    	\includegraphics[width=0.8\linewidth]{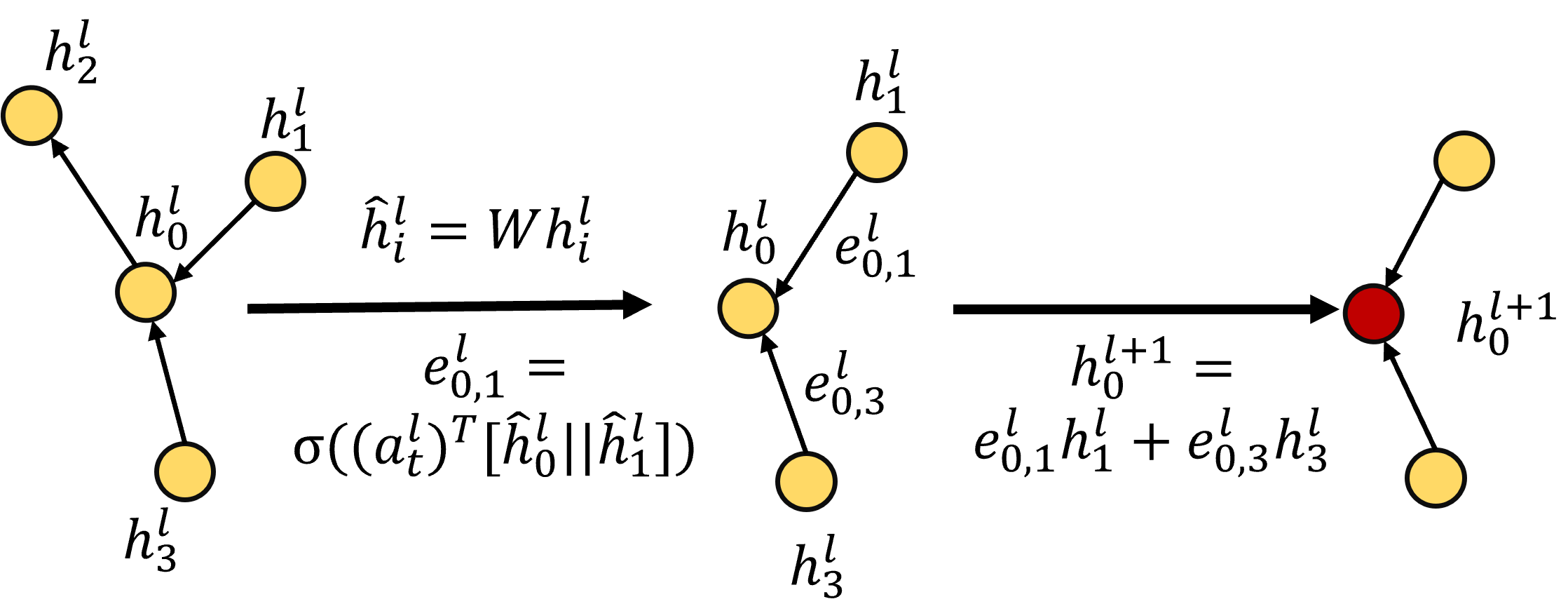}   %以pic.jpg的0.5倍大小输出
    	\label{fig:gat}
    	%\end{minipage}
    } \\
    \subfigure[DGAT]{
    	%\begin{minipage}[t]{0.14\textwidth}
    	%\centering          %子图居中
    	\includegraphics[width=0.95\linewidth]{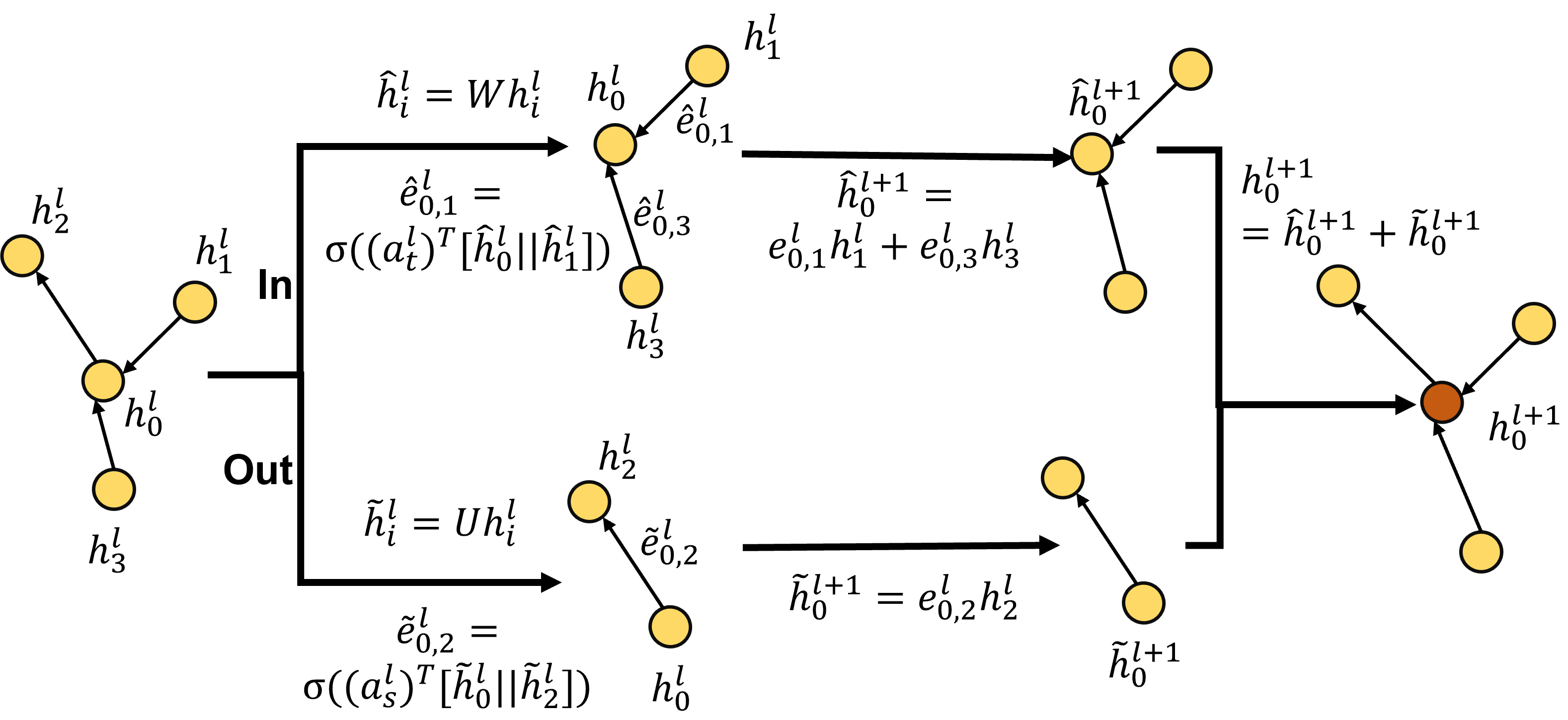}   %以pic.jpg的0.5倍大小输出
    	\label{fig:dgat}
    	%\end{minipage}
    } \\
    \subfigure[DEDGAT]{
    	%%\begin{minipage}[t]{0.14\textwidth}
    	%\centering          %子图居中
    	\includegraphics[width=1.0\linewidth]{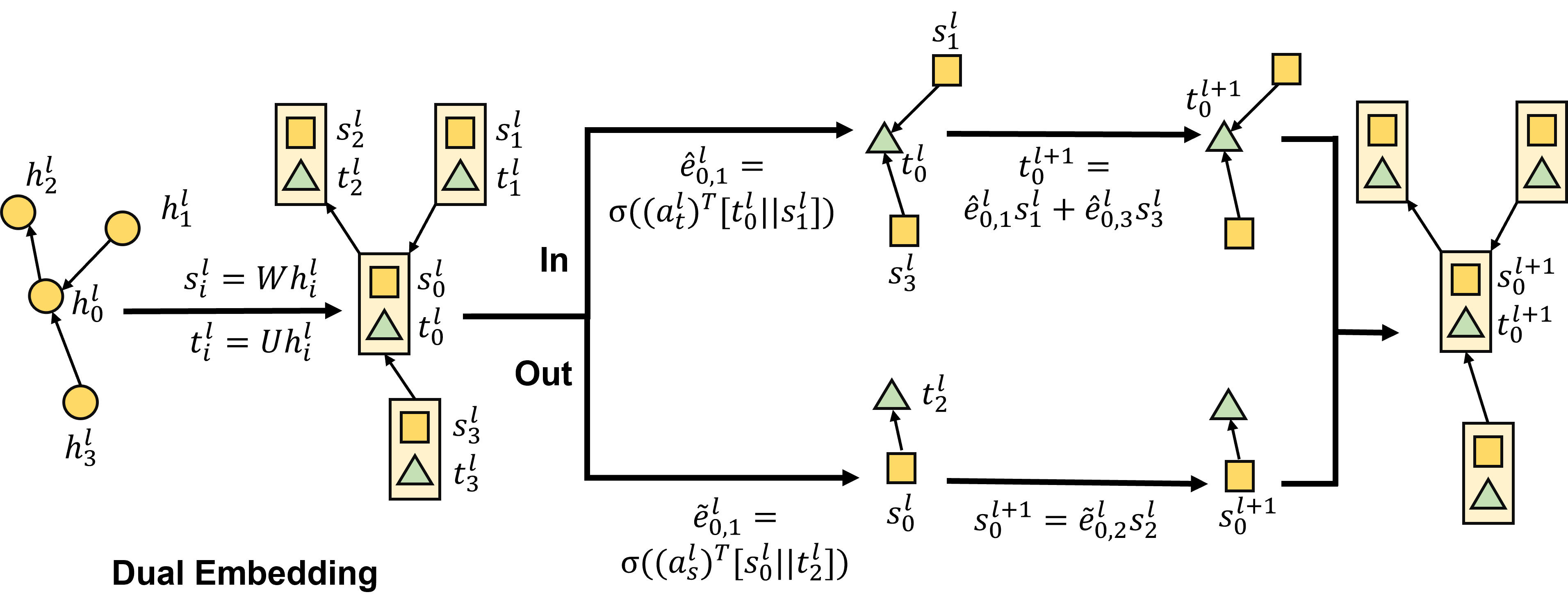}   %以pic.jpg的0.5倍大小输出
    	\label{fig:dedgat}
    	%\end{minipage}
    }
  \caption{Different Aggregation processes. (a) GAT only aggregate in-degree features in one layer. (b) DGAT utilize both in-degree and out-degree features. (c) DEDGAT seperates a node to in-degree (triangles) and out-degree (squares)  nodes and computes two different embeddings for in- and out-degree feature represntations, respectively.}
  \label{fig:aggPro}
\end{figure}

As shown in Fig.~\ref{fig:gat}, the undirected GAT can only aggregate in-degree information, resulting in insufficient out-degree information. This leads to the problem of directional bias as shown in Fig.~\ref{fig:weight}.
To explicitly exploit directed information of graph nodes, we propose a directed graph attention network (DGAT) to calculate both in-degree and out-degree attentions, respectively, as shown in Fig.~\ref{fig:dgat}. To further distinguish in-degree and out-degree informaton, we propose a dual embedding in terms of DGAT
(as referred to DEDGAT) to extract discriminative features for different directions of graph nodes, as shown in Fig.~\ref{fig:dedgat}.
%%两句话描述一下GAT以及缺点，引出dgat和degat
\label{sec:typestyle}
% \subsection{Directed Attention Mechanism}
% \begin{figure}[h]
%   \centering
%   \includegraphics[width=0.5\textwidth]{imgs/da2.PNG}
%   \caption{Directed attention for a single step of propagation.}
%   \label{fig:da}
% \end{figure}
% 为了提取节点的有向拓扑信息,本文提出有向注意机制(称为$ DA $)分别计算节点的入度和出度的注意力,如图.~\ref{fig:dade}所示。
\subsection{Directed Graph Attention Networks (DGAT)}

% 给定第 $l$ 层节点 $i$ 的表示 $h_i^{(l)}$,其邻居 $j_1,j_2,\cdots$ 的表示为 $h_{j_1}^{(l )},h_{j_2}^{(l)},\cdots$,传播前的线性插值是$\hat{h}_i^{(l)}=W^{(l)}h_i^{(l )}$ 和 $\tilde{h}_i^{(l)}=U^{(l)}h_i^{(l)}$ 分别用于邻居和邻居。
Given the representation $h_i^{(l)}$ of the node $i$ at the layer $l$, the representation of its neighbors $j_1, j_2, \cdots$ is $h_{j_1}^{(l )}, h_{ j_2}^{(l)}, \cdots$, and the linear transformation before propagation is given by $\hat{h}_i^{(l)}=W^{(l)}h_i^{(l )}$ and $ \tilde{h}_i^{(l)}=U^{(l)}h_i^{(l)}$ for in-degree and out-degree respectively.
Then, two map functions
$\mathcal{F}:\hat{h}\in\mathbb{R}^{2d}\rightarrow \hat{e}\in\mathbb{R}^{1}$
and
$\mathcal{G}:\tilde{h}\in\mathbb{R}^{2d}\rightarrow \tilde{e}\in\mathbb{R}^{1}$
are used to calculate the attention weights of the neighbors as follows,
\begin{align} \hat{e}_{i,j}^{(l)}&=\sigma((a_t^{(l)})^{T}[\hat{h}_i^{(l)} \| \hat{h}_j^{(l)}]),
    \label{equ:equ1}
\end{align}
\begin{align}
    \tilde{e}_{i,j}^{(l)}&=\sigma((a_s^{(l)})^{T}[\tilde{h}_i^{(l)} \| \tilde{h}_j^{(l)}]),
\label{equ:equ2}
\end{align}
where $\|$ represents the concatenation operator, $\sigma$ represents the softmax function, and $a_t^{(l)}$ and $a_s^{(l)}$ are the parameters of the inner and outer neighbor attention mechanisms.
% 在对邻居的注意力权重进行softmax之后,节点 $i$ 在层 $l+1$ 的表示是
After softmaxing the attention weights of neighbors,
the representation of node $i$ at level $l+1$ is given by Eq.~\ref{equ:equ3}, where $\mathcal{T }(i)$ and $\mathcal{S}(i)$ represent the set of inner and outer neighbors, respectively.
\begin{align}
h_i^{(l+1)}=&\hat{h}_i^\text{'(l)}+\tilde{h}_i^\text{'(l)}  \notag\\
	 =& \sum_{j\in\mathcal{T}(i)}\hat{e}_{i,j}^{(l)}\hat{h}_j^{(l)} + \sum_{j\in\mathcal{S}(i)}\tilde{e}_{i,j}^{(l)}\tilde{h}_j^{(l)}  \notag\\
	 =&\sum_{j\in\mathcal{T}(i)}\hat{e}_{i,j}^{(l)}W^{(l)}h_j^{(l)} + \sum_{j\in\mathcal{S}(i)}\tilde{e}_{i,j}^{(l)}U^{(l)}h_j^{(l)}.
\label{equ:equ3}
\end{align}

\subsection{Dual Embedding of DGAT (DEDGAT)}
DGAT adequately adopts in-degree and out-degree information. Nevertheless, the directivity of feature is still not taken into account. We argue that single feature cannot fully reflect the difference between input and output information.
Therefore, we propose a directed attention with dual embedding model DEDGAT.
As shown in Fig.~\ref{fig:dedgat}
, each node consists of two representations $t_i^{(l)}$ and $s_i^{(l)}$ to extract node features of in-degree and out-degree, respectively.
Before computing attention, there are two linear transformation defined as $ \hat{t}_i^{(l)} = w^{(l)}t_i^{(l)} $ and $ \hat{s}{ (l)} = u^{(l)}s_i^{(l)}$ for $\quad i= 1, 2, \cdots, n $
where $w^{(l)}$ and $u^{(l)}$ are the weights of linear layer. Second, the attention weights of in-degree and out-degree are calculated based on Eq.(\ref{equ:equ4}-\ref{equ:equ6}), where $a_t^{(l)}$ and $a_s^{(l)}$ are the parameters of the inner and outer neighbor attention mechanisms.
\begin{align}
	 \hat{e}_{i,j}^{(l)}&=\sigma((a_t^{(l)})^{T}[\hat{t}_i^{(l)}||\hat{s}_j^{(l)}]),
\label{equ:equ4} \\
%\end{align}
%\begin{align}	 
 \tilde{e}_{i,j}^{(l)}&=\sigma((a_s^{(l)})^{T}[\hat{s}_i^{(l)}||\hat{t}_j^{(l)}]).
	\label{equ:equ6}
\end{align}
Then, the two branches of the in-degree and out-degree attentions are aggregated respectively to generate two representations for the next layer.
On the last layer $l$, the in-degree representation $t_i^{(l)}$ and the out-degree representation $s_i^{(l)}$ are added to generate the representation of node $i$. The added features are used to predict the network output.

\subsection{Node Classification}
We use two DEDGAT layer following a linear layer to conduct node classification on three financial graph datasets. Given the embedding generated by the last DEDGAT layer $L$, the prediction for node classification is given by:
\begin{align}
	\hat{y}_{i,c}=\text{softmax}_c(Vh_{i}^{(L)}),
	\label{equ:node_classification}
\end{align}
where $V$ denote the final linear layer.

\section{Experiments}
\label{sec:exp}

\subsection{Datasets}
We evaluate our approach on three directed graph datasets. Among them, the Elliptic dataset\cite{weber2019anti} is to detect money laundering from transaction data, the Ant dataset is to detect financial fraud from transfer records, the DGraphFin dataset~\cite{huang2022dgraph} is to detect financial fraud from social networks.
The Ant dataset is self-collected and others are open-sourced. Each dataset contains two categories, 0 represents normal nodes and 1 represents abnormal nodes. Optimizing node classification tasks on these datasets can help us detect financial risk.
Table.~\ref{table:dataset-gmlat} shows the statistics of three datasets.
\begin{table}[htb]
	\centering
	\caption{Financial Risk Control Datasets}
	\begin{tabular}{lccc} %第一列居中显示,第二列左对齐
		\toprule
		- & Elliptic & Ant & DGraphFin \\
		\midrule
		Nodes & 203,769 & 1,296,486  & 3,700,550 \\
		Edges & 234,355 & 1,785,844 & 4,300,999 \\
		Labeled Nodes & 203,769 & 1,296,486 & 1,225,601 \\
		Features & 93 & 59 & 17 \\
		Classes & 2 & 2 & 2 \\
		\bottomrule
	\end{tabular}
	\label{table:dataset-gmlat}
\end{table}
\subsection{Baselines and Parameter Settings}
We select GAT\cite{velivckovic2017graph}, GCN\cite{kipf2016semi}, GraphSage\cite{hamilton2017inductive}, DGCN\cite{ma2019spectral}, MagNet\cite{zhang2021magnet} methods for comparison. Among them, GAT, GCN and GraphSage are basic graph models, DGCN, MagNet are latest graph models designed for directed graphs.

To compare openly and transparently, we provide Tab.~\ref{tab:net_parameters} which shows the network parameters of each GNN model used in our experiments.
\begin{table}[htb]
\centering
\caption{Numbers of Network Parameters}
%\resizebox{0.5\textwidth}{2cm}{
\begin{tabular}{crrc}
\toprule
        & Elliptic & Ant     & DGraphFin \\
       \midrule
GCN            & 98,306   & 63,490  & 20,482    \\
GAT             & 113,696  & 78,880  & 35,872    \\
SAGE            & 98,818   & 64,002  & 20,094     \\
DGCN            & 620,82   & 57,730  & 52,354    \\
MagNet          & 134,917  & 121,861 & 89,842   \\
\midrule
DGAT              & 57,892   & 40,484  & 89,160   \\
DEDGAT            & 114,756  & 79,940  & 88,200    \\
\bottomrule
\end{tabular}
%}
\label{tab:net_parameters}
\end{table}

\subsection{Implementations}
We use AGL~\cite{aglpaper}, a scalable, fault-tolerance and integrated system, with fully-functional training and inference for GNNs to train our networks. The AGL system consists of three parts, i.e., GraphFlat, GraphTrainer, GraphInfer, respectively. GraphFlat is an efficient and distributed generator based on message passing to generate K-hop
neighborhoods that contains information complete subgraphs of each targeted nodes. GraphTrainer combines pipeline, pruning, and edge-partition modules to alleviate the overhead on I/O and optimize the floating point calculations during the training of GNN models. GraphInfer is a distributed inference module that splits K layer GNN models into K slices, and applies the message passing K times based on MapReduce. The three computing logics above are built on the distributed file system, CPU cluster, and MapReduce computing framework. Our AGL system, implemented on mature infrastructures, can finish the training of a 2-layer graph attention network on a directed graph with billions of nodes and hundred billions of edges in 14 hours.

\subsection{Classification performance}
Table.~\ref{tab:all_results} shows the experimental results of each method on the node classification task. Our DEDGAT effectively improves the classification performance, achieves the best results on all three datasets.
\begin{table}[htb]
\centering
\caption{Node classification performance(AUC)}
%\resizebox{0.5\textwidth}{2cm}{
\begin{tabular}{cccc}
\toprule
       & Elliptic & Ant            & DGraphFin           \\
       \midrule
GCN    & 91.09±0.28 & 85.18±0.57 & 70.78±0.23     \\
GAT    & 95.06±0.45 & 97.43±0.24 & 76.24±0.81     \\
SAGE   & 95.79±0.33 & 97.49±0.19 & 77.61±0.18 \\
DGCN   & 94.48±0.60 & 94.71±0.85 & 79.58±0.15 \\
MagNet & 92.84±1.01 & 91.95±0.20 & 79.71±0.27  \\
\midrule
DGAT     & 95.79±0.43 & 97.92±0.31 & 80.72±0.11 \\
DEDGAT   & \textbf{96.18±0.23} & \textbf{98.16±0.09} & \textbf{81.37±0.06}   \\
\bottomrule
\end{tabular}
%}
\label{tab:all_results}
\end{table}

% \begin{figure*}[htb]
% 	\centering
%  %{\includegraphics[height=2.2cm,width=8.5cm]{imgs/pge_imgs/4938_new.png}\label{fig:whydirected-in-gt1}}
%  {\includegraphics[width=\textwidth]{imgs/7102.png}}
% 	\caption{7102.}
% 	\label{fig:7102}
% \end{figure*}

\begin{figure*}[htb]
\begin{minipage}[]{0.31\textwidth}
    \centering
    \centerline{\includegraphics[width=5.5cm]{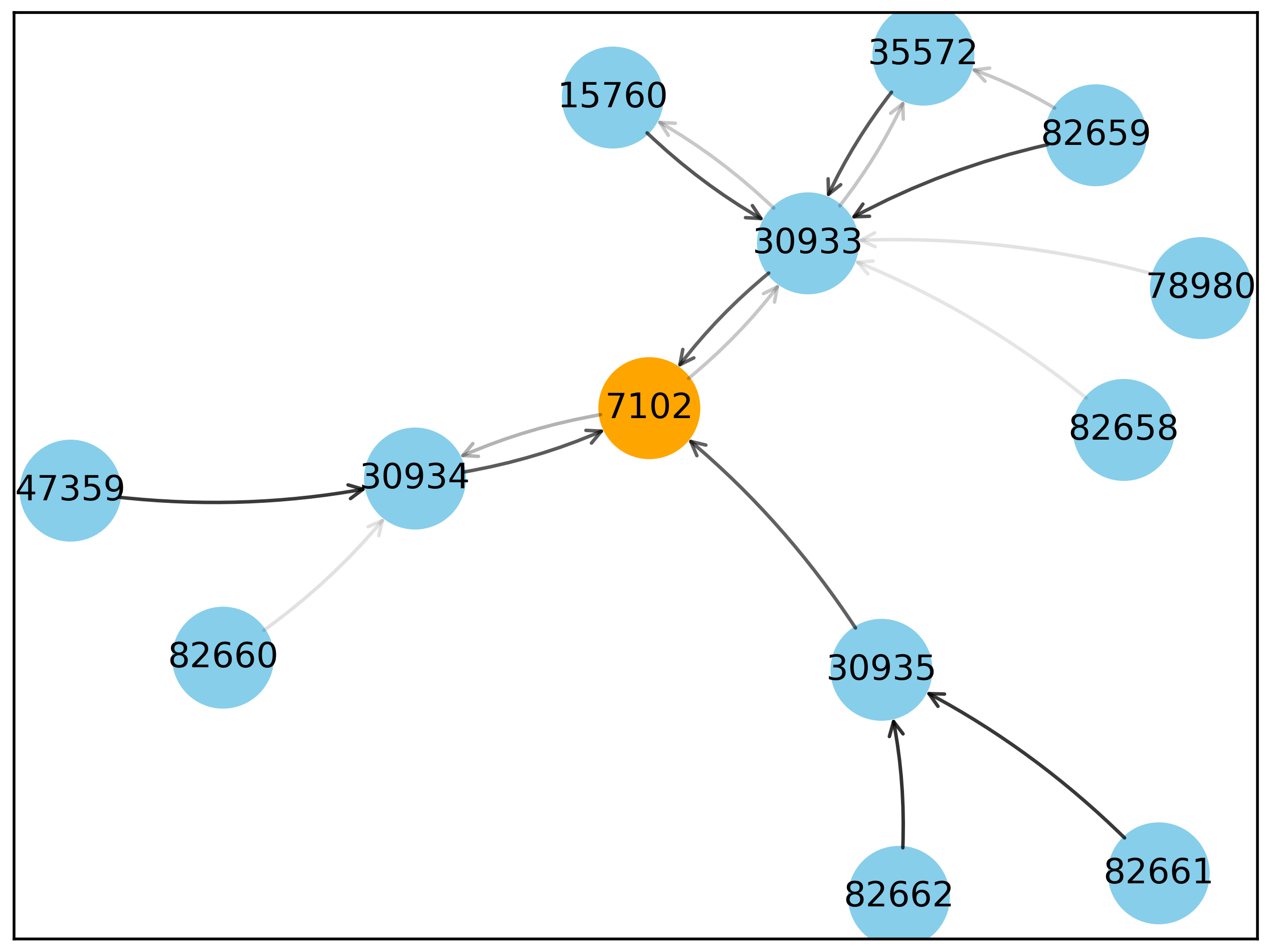}}
    \centerline{(a) GAT}
\end{minipage}
\begin{minipage}[]{0.31\textwidth}
    \centering
    \centerline{\includegraphics[width=5.5cm]{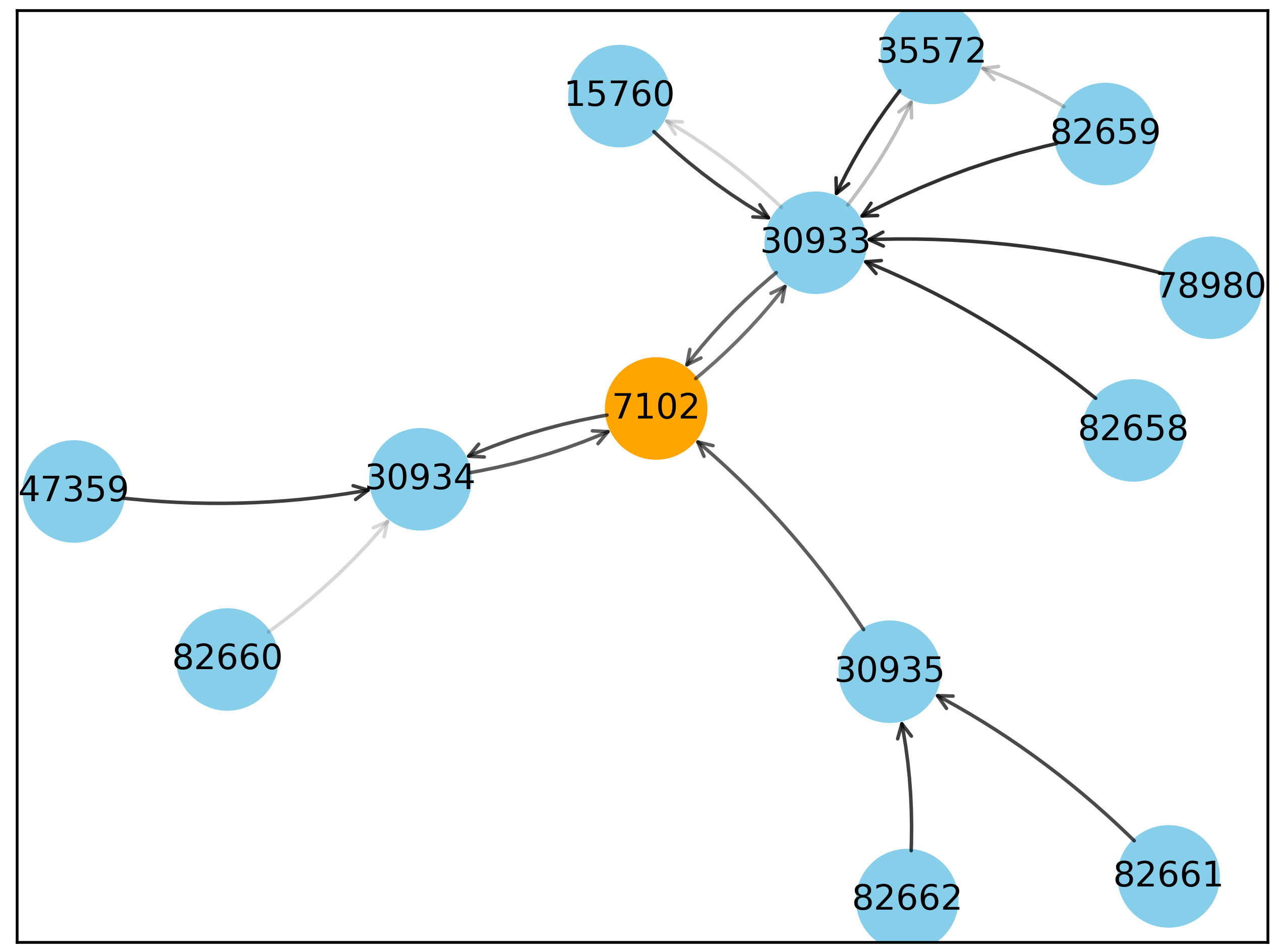}}
    \centerline{(b) DGAT}
\end{minipage}
\begin{minipage}[]{0.31\textwidth}
    \centering
    \centerline{\includegraphics[width=5.5cm]{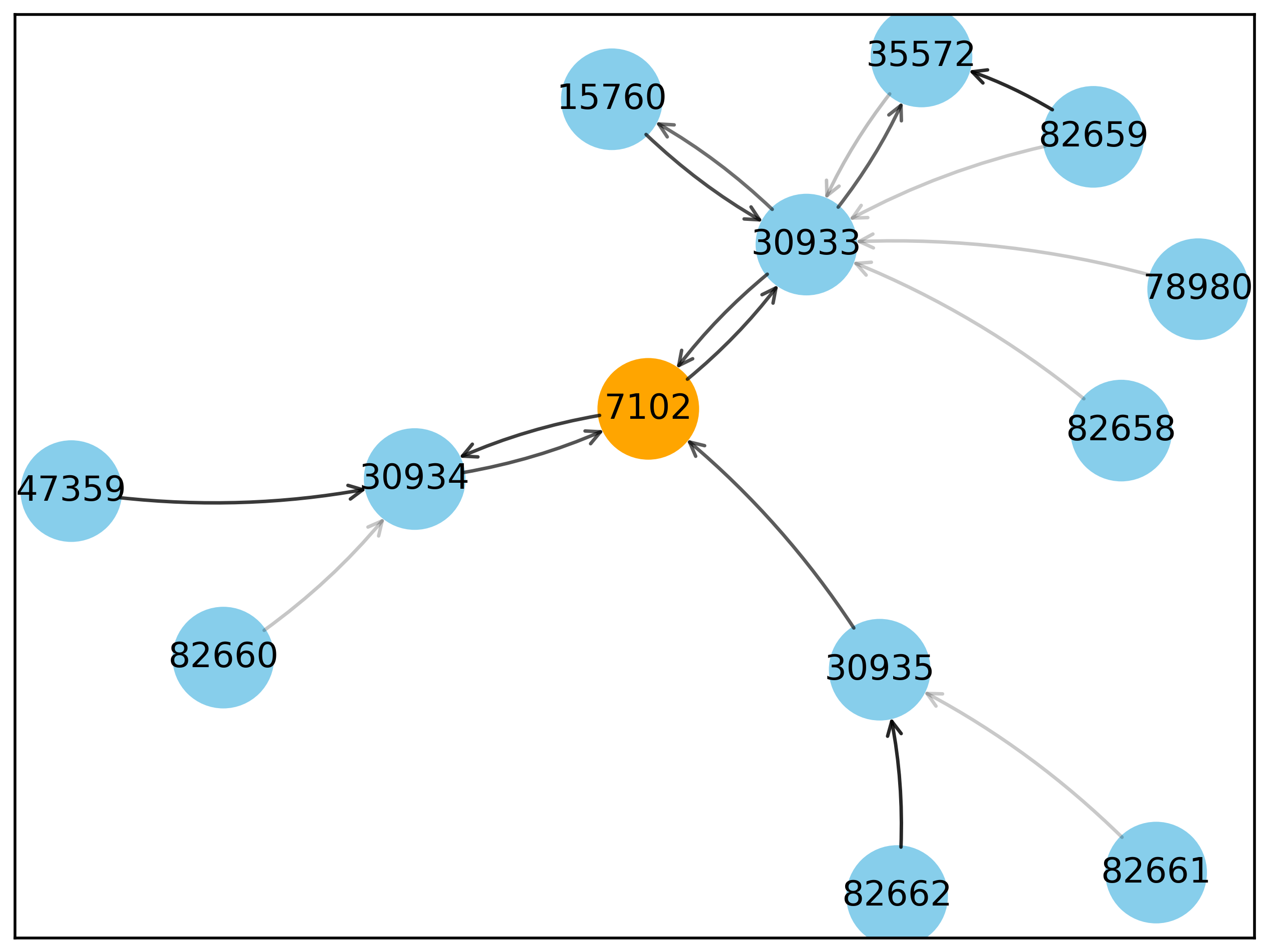}}
    \centerline{(c) DEDGAT}
\end{minipage}

\caption{Comparison of interpretation of target node 7102 by GNNExplainer. GAT ignores four out-degree edges $($7102 $\to$ 30934$)$, $($ 7102 $\to$ 30933 $)$, $($ 30933 $\to$ 1576$)$ and $($ 30933 $\to$ 35572$)$, DGAT resorts to two out-degree edges $($7102 $\to$ 30934$)$ and $($ 7102 $\to$ 30933 $)$, and DEDGAT leverages all four edges which are ignored by GAT.}
\label{fig:7102}
\end{figure*}
\subsection{Visualization results}
%介绍 explianer，以及实现方法
\label{sec:exper}

When GAT~\cite{velivckovic2017graph} is applied to undirected graphs, the nodes only aggregate information from the in-degree direction. In a directed graph, edge ${E_{ij}}$ represent the out-degree edge of node i and the in-degree edge of node j. GAT implicitly uses out-degree information based on the directional symmetry. In the directional context of digraph, GAT-like methods which implicitly fuse out-degree information cannot generalize well to the recognition scene of directed graphs. However, Our method explicitly separates the out-degree and in-degree information for calculation, and introduces dual embedding in DEDGAT to emphasize the idea of separation. The purpose of discrete modeling is to alleviate the in-degree bias that exists in GAT.

GNNExplainer~\cite{ying2019gnnexplainer} is a graph interpreter, it outputs the importance score of each edge in graph model. To verify the validity of the directionality we introduced, we use GNNExplainer to visualize the aggregation process of three methods : GAT, DGAT and DEDGAT. As Fig.~\ref{fig:7102} shows (Darker edges mean higher importance): GAT underutilizes the four out-degree edges $($7102 $\to$ 30934$)$, $($ 7102 $\to$ 30933 $)$, $($ 30933 $\to$ 1576$)$ and $($ 30933 $\to$ 35572$)$; In contrast, our DGAT resorts to two out-degree edges $($7102 $\to$ 30934$)$ and $($ 7102 $\to$ 30933 $)$ ; DEDGAT steps futher and leverages all four edges which ignored by GAT.

We traverse the Ant dataset to provide global statistics result(Fig.~\ref{fig:weight}) for ${S_{in}}$ and ${S_{out}}$,where ${S_{in}}$ and ${S_{out}}$ are defined as follows:
\begin{equation}
    	\begin{gathered}
  {S_{in}} = \left\{ {{w_{ij}} \in \left. S \right|k\left( i \right) > k\left( j \right)} \right\}, \hfill \\
  {S_{out}} = \left\{ {{w_{ij}} \in \left. S \right|k\left( i \right) < k\left( j \right)} \right\}, \hfill \\ 
\end{gathered} 
    	\label{equ:equ8}
\end{equation}
where $k\left( i \right)$ represents the number of hops from node i to the 
central risk node. S is defined as $S = \left\{ {\left. {{w_{ij}}} \right|I\left( {i,j} \right) = 1} \right\}$, where ${w_{ij}}$ is from GNNExplainer, i.e.,
${w_{ij}} = GNNExplr\left( {{E_{ij}}} \right), {E_{ij}} \in G\left( {E,V} \right)$. Here, $I\left( {i,j} \right)$ is the indicative function 
\begin{equation}
    	I\left( {i,j} \right) = \left\{ {\begin{array}{*{20}{c}}
  {1,{A_{ij}} = 1 \wedge {A_{ji}} = 1}, \\ 
  {0,{A_{ij}} = 0 \vee {A_{ji}} = 0} ,
\end{array}} \right.
    	\label{equ:equ12}
\end{equation}
where A is the adjacency matrix of the directed graph G.

\begin{figure}[htb]
\begin{minipage}[]{0.13\textwidth}
    \centering
    \centerline{\includegraphics[width=2cm]{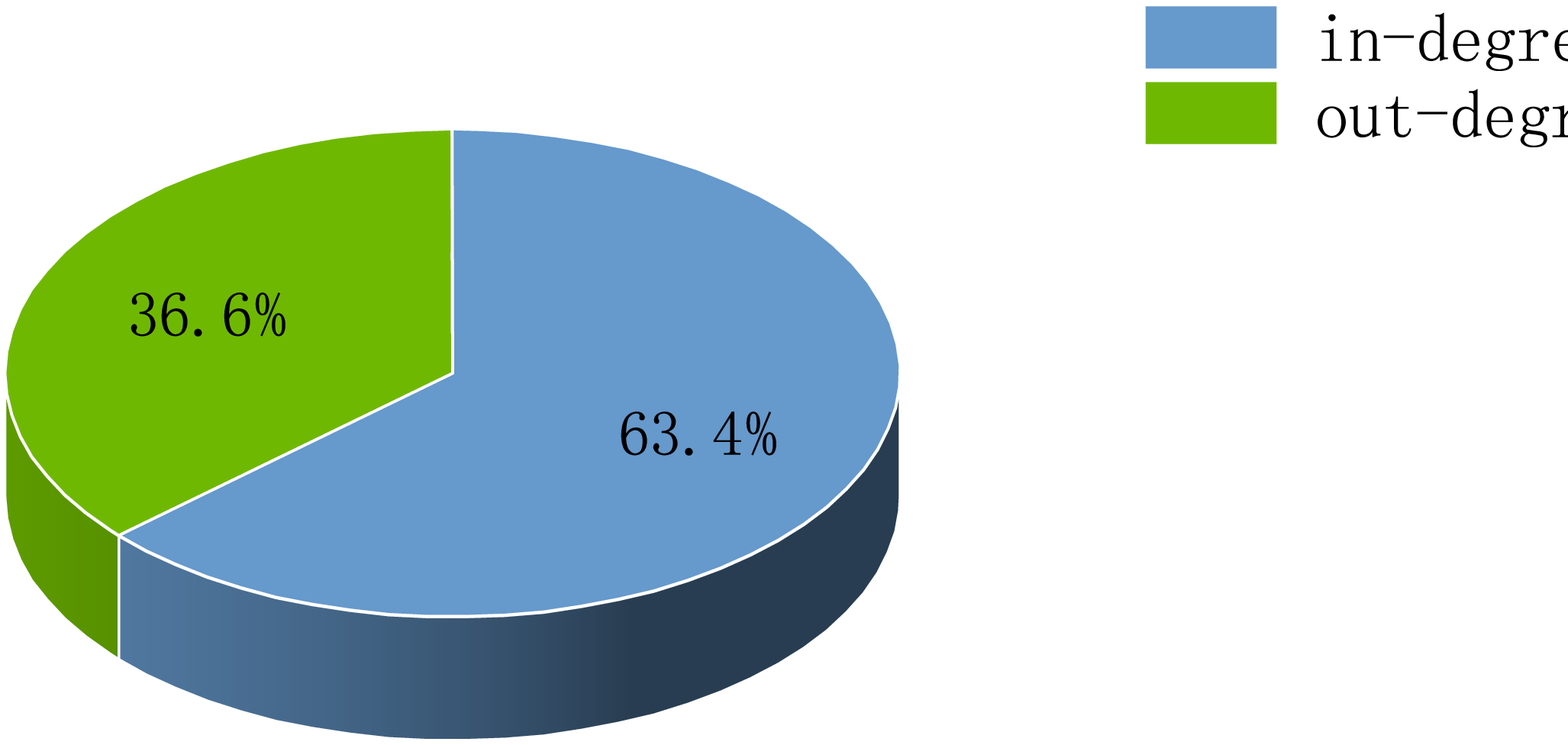}}
    \centerline{(a) GAT}
\end{minipage}
\begin{minipage}[]{0.15\textwidth}
    \centering
    \centerline{\includegraphics[width=2cm]{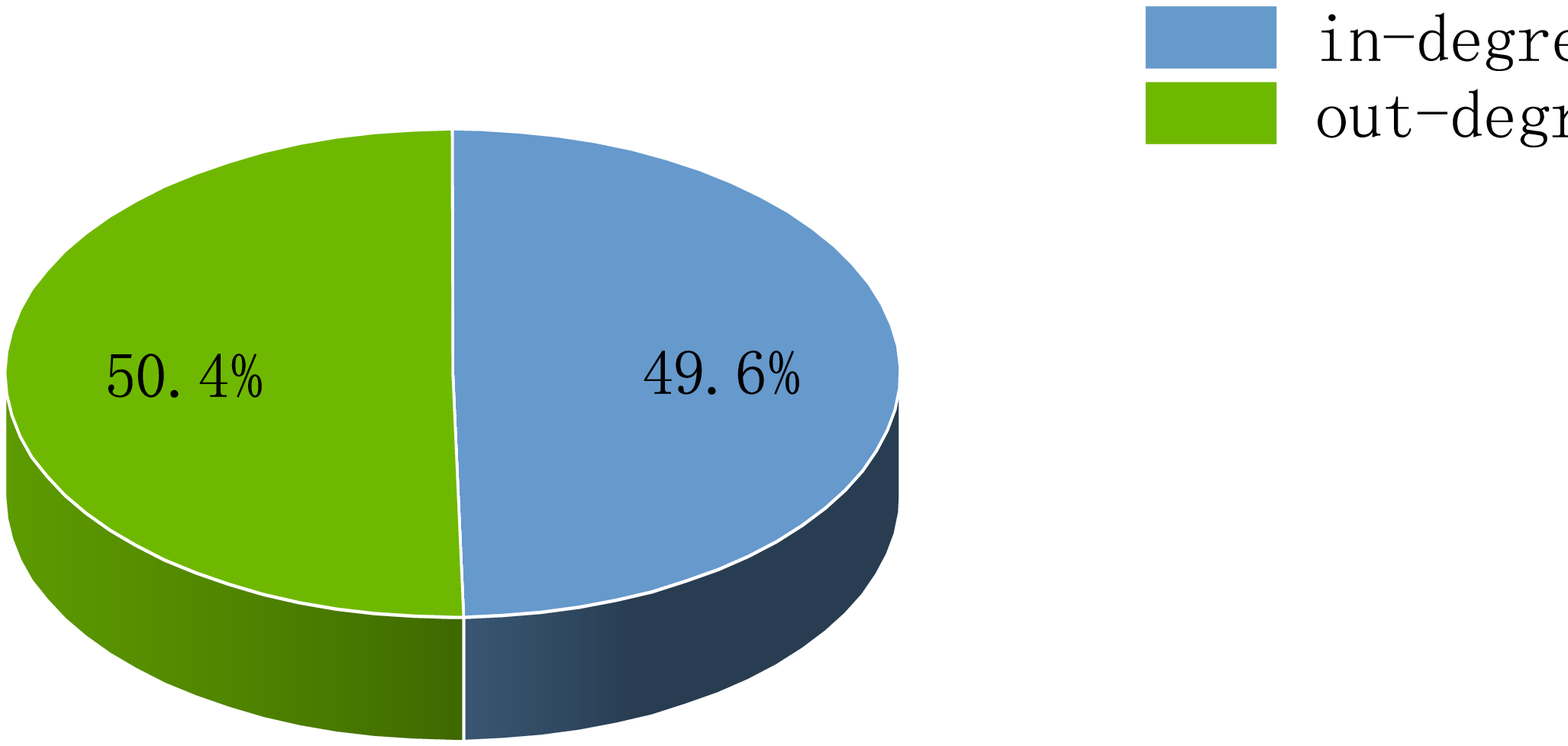}}
    \centerline{(b) DGAT}
\end{minipage}
\begin{minipage}[]{0.15\textwidth}
    \centering
    \centerline{\includegraphics[width=2.5cm]{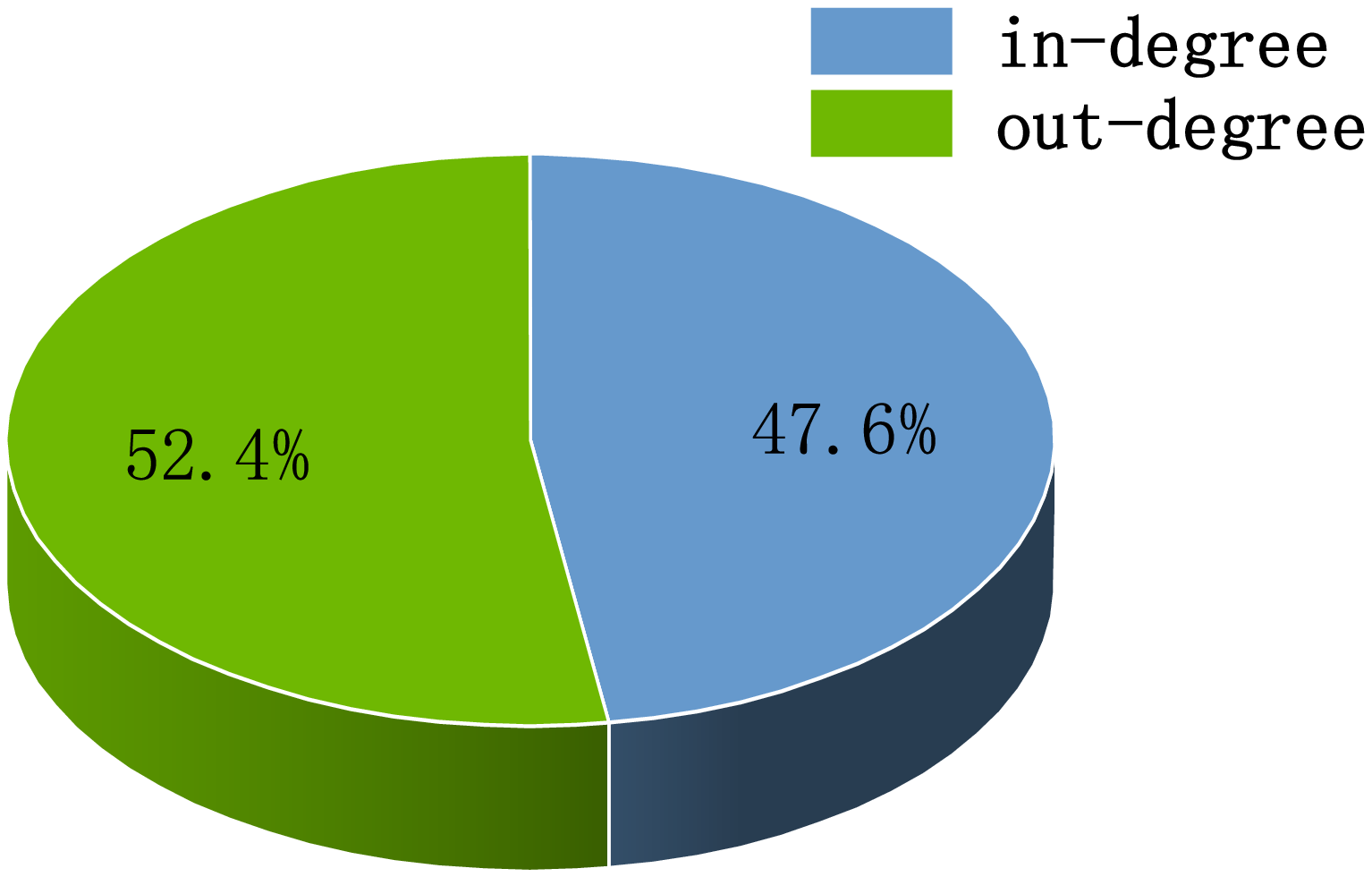}}
    \centerline{(c) DEDGAT}
\end{minipage}

\caption{The proportion of in-degree and out-degree edge weights when their importance scores are larger than 0.5. The proportion of the in-degree edge weights of GAT is 1.73 times than that of the out-degree ones, while the proportion of the two directional edge weights in the proposed DGAT and DEDGAT is almost the same.}
\label{fig:wight}
\end{figure}

As shown in Fig.~\ref{fig:weight}, GAT relies excessively on the in-degree information, but out-degree edges are under exploited;
Benefiting from DGAT and DEDGAT's separate calculation of in-degree and out-degree, both in-edges and out-edges are assigned approximate weights. We set threshold $\lambda $ to 0.5 and count those in-degree and out-degree edges whose weights are larger than the threshold(these edges are important in the graph convolution). Statistics results are shown in Fig.~\ref{fig:wight}: 
GAT emerges a sharp in-degree information bias. But DGAT and DEGAT keep the in-degree and out-degree information on a balance level.

%\subsection{Implementations}
%We use AGL~\cite{aglpaper}, a scalable, fault-tolerance and integrated system, with fully-functional training and inference for GNNs to train our networks. As Fig.~\ref{fig:agl} shows, the AGL system consists of three parts named GraphFlat, GraphTrainer, GraphInfer. GraphFlat is an efficient and distributedgenerator, based on message passing, for generating K-hop neighborhoods that contain information complete subgraphs of each targeted nodes. GraphTrainer combines pipeline, pruning, and edge-partition module to alleviate the overhead on I/O and optimize the floating point calculations during the training of GNN models. GraphInfer is a distributed inference module that splits K layer GNN models into K slices, and applies the message passing K times based on MapReduce. The three computing logics above are built on the distributed file system, CPU cluster, and MapReduce computing framework. Our system AGL, implemented on mature infrastructures, can finish the training of a 2-layer graph attention network on a graph with billions of nodes and hundred billions of edges in 14 hours.

%\begin{figure}[htb]
	%\centering
 %{\includegraphics[height=2.2cm,width=8.5cm]{imgs/pge_imgs/4938_new.png}\label{fig:whydirected-in-gt1}}
 %{\includegraphics[width=0.5\textwidth]{imgs/agl.png}}
	%\caption{Our AGL Architecture.}
	%\label{fig:agl}
%\end{figure}

\section{Conclusion}
\label{sec:prior}
In this paper, we have proposed a directed graph attention network (referred to as DGAT) to explicitly calculate in-degree and out-degree representation of a directed graph to tackle the problem of directional bias, such as the weight important distribution shown in Fig.~\ref{fig:gat} for graph attention network (GAT)~\cite{velivckovic2017graph}.
In order to better distinguish the in-degree and out-degree node representation of the directed graph, we split each node into in-degree and out-degree nodes. A dual embedding of DGAT model (named as DEDGAT) is further proposed to separately calculate the node representations of the incoming and outgoing degrees, respectively. Both DGAT and DEDGAT can effectively solve the problem of the directional bias of GAT. Also, DEDGAT achieves the best performance on the three financial risk control datasets. Visualization results by GNNExplainer~\cite{ying2019gnnexplainer} demonstrate the importance of directionality for financial risk control tasks. %However, DEDGAT only aggregates the node features, and ignores edge features in the financial risk control dataset. Therefore, follow-up research will further explore how to utilize edge features to help financial risk control.

% In this paper, we have proposed a novel graph attention network DEDGAT

% Identifying abnormal nodes in the financial relationship network is of great significance in the field of financial risk control. The transfer in and transfer out in financial relationship networks have different information, our experiments verify that the undirected graph model GAT is in-degree biased and cannot fully consider information in both directions. So, we design the directed attention with dual embedding model (DEDGAT) to alleviate the bias. Experiments show that DEDGAT achieves SOTA performance on three financial datasets, which proves the importance of bidirectional information in the field of financial risk control.
\vfill\pagebreak
%\bibliographystyle{ACM-Reference-Format}
%\bibliography{sample-base}

 % <============================================= mwe.bbl
\end{document}